\documentclass{eptcs}
\providecommand{\event}{Post-proceedings of DCM 2015}

\usepackage{macros}
\usepackage{pdfpages}
\usepackage[all]{xypic}

\begin{document}

\newcommand{\DAG}{\D}
\newcommand{\actions}{\mathcal{R}}
\newcommand{\godown}{\mathbin{\D^\star}}
\renewcommand{\edges}{\mathbin{\E}}
\newcommand{\edgesR}{\mathbin{\overrightarrow{\E}}}
\newcommand{\edgesL}{\mathbin{\overleftarrow{\E}}}
\newcommand{\MG}{\mathbf{MG}}
\newcommand{\MSG}{\mathbf{MSG}}

\title{A knowledge representation meta-model for rule-based modelling of signalling networks}

\author{
Adrien Basso-Blandin
\institute{LIP, ENS Lyon,\\Lyon, France}
\email{adrien.basso-blandin@ens-lyon.fr}
\and
Walter Fontana
\institute{Harvard Medical School,\\Boston, USA}
\email{walter@hms.harvard.edu}
\and
Russ Harmer
\institute{CNRS \& LIP, ENS Lyon,\\Lyon, France}
\email{russell.harmer@ens-lyon.fr}
}

\def\titlerunning{A KR meta-model for rule-based modelling}
\def\authorrunning{Basso-Blandin et al.}

\maketitle

\thispagestyle{empty}
\pagestyle{empty}

\begin{abstract}

The study of cellular signalling pathways and their deregulation in disease states, such as cancer, is a large and extremely complex task. Indeed, these systems involve many parts and processes but are studied piecewise and their literatures and data are consequently fragmented, distributed and sometimes---at least apparently---inconsistent. This makes it extremely difficult to build significant explanatory models with the result that effects in these systems that are brought about by many interacting factors are poorly understood.

The rule-based approach to modelling has shown some promise for the representation of the highly combinatorial systems typically found in signalling where many of the proteins are composed of multiple binding domains, capable of simultaneous interactions, and/or peptide motifs controlled by post-translational modifications. However, the rule-based approach requires highly detailed information about the precise conditions for each and every interaction which is rarely available from any one single source. Rather, these conditions must be painstakingly inferred and curated, by hand, from information contained in many papers---each of which contains only part of the story.

In this paper, we introduce a graph-based meta-model, attuned to the representation of cellular signalling networks, which aims to ease this massive cognitive burden on the rule-based curation process. This meta-model is a generalization of that used by Kappa and BNGL which allows for the flexible representation of knowledge at various levels of granularity. In particular, it allows us to deal with information which has either too little, or too much, detail with respect to the strict rule-based meta-model. Our approach provides a basis for the gradual aggregation of fragmented biological knowledge extracted from the literature into an instance of the meta-model from which we can define an automated translation into executable Kappa programs.

\end{abstract}

\section{Introduction}\label{intro}

We propose a knowledge representation (KR) meta-model to enable the study of the \emph{dynamics} of cellular signalling networks and, in particular, the consequences of mutations on dynamics. Our aim is therefore not to construct a Description Logic-based terminology (or any other ontology of this general kind) of static concepts to perform inference of the kind ``ERK is an enzyme that phosphorylates so it is a kinase"; nor is it to build a representation \emph{of} dynamics to support inference about the time-evolution of systems. Rather, we seek to represent each individual protein-protein interaction (PPI) that constitutes a signalling network as a formal \emph{rule} that expresses the known, empirically necessary conditions for that PPI to occur.
These rules resemble those of Kappa or BNGL \cite{hlavacek2006rules} but need not respect the stringent meta-model imposed by those formalisms wherein all bonds must occur between explicitly specified sites and all other relevant factors---such as protein conformation, post-translational modifications (PTMs) or, more generally, the presence or absence of key residues---are opaquely encoded into monolithic states attached to sites. These rules can then be automatically assembled into \emph{bona fide} rule-based [Kappa] models that can be simulated, subjected to static analysis and whose causal structure can be examined in detail.

The need for such a two-phase approach to the rule-based modelling of signalling networks
arises because the pertinent information is dispersed across the literature in such a way that any given paper typically contains fragments, or \emph{nuggets}, of \emph{partial} mechanistic knowledge about multiple PPIs; and nuggets appear, for any given PPI, in many papers. As such, the basic currency of the first phase, \ie the \emph{curation} process, must be nuggets, corresponding to rules in our generalized meta-model, as opposed to rules in the strict rule-based meta-model; and the curation process focusses not directly on the PPIs themselves---as would necessarily the direct curation of rules---but rather on extracting nuggets, identifying which PPIs they refer to and incrementally aggregating them into more detailed nuggets. This enables a model-agnostic curation process that avoids hard-wiring invariants that might be true in one context, or model, but not another, \eg in one particular model, two physically-overlapping binding sites of a protein could be represented by a single formal site; but, in the open world of evolving curated knowledge, they must be kept formally separate as they may need to be distinguishable in some model(s), typically if a third physical binding site is later discovered that overlaps with one but not the other.

The second \emph{instantiation} phase of our model building process concerns the passage from curated knowledge to a purely formal representation. In this step, we produce---from a \emph{given} state of knowledge---a strict rule-based model where invariants may be hard-wired, if they are warranted by that knowledge. This phase begins with the specification of which proteins are to be represented and then determines (i) which nuggets apply to those proteins; and (ii) all \emph{site conflict} invariants implied by the knowledge in question. The first point is important because our notion of agent in the KR represents not a single gene product but a \emph{neighbourhood in sequence space}. This design decision is central to our approach and enables us to represent interactions that depend on certain key residues in such a way that the effects of both loss- and gain-of-function mutations can be automatically determined by inspecting whether or not a particular protein has the appropriate key residues. This resolves a dangling question from earlier work on `meta Kappa' \cite{danos2009rule,harmer2009rule} which was only able to capture the effects of loss-, but not gain-, of-function mutations. The second point is critical in order that the formal representation match the physical constraints, present in the curated knowledge but necessarily left open in the KR---as discussed above.

It is instructive to compare our approach with that embodied by production rule-based expert systems, such as MYCIN \cite{buchanan1984rule} and its descendants, which saw a gradual drift away from considering each rule as an independent element [as we do] towards a view where the KR should be ``designed and maintained as a coherent whole" \cite{clancey1993notes}. Such a viewpoint may be entirely appropriate in a domain where human experts can be reasonably expected to agree on most points; but, in a domain characterized by a large, dispersed and fragmented body of knowledge that no single human expert can hope to master, expert opinion cannot reasonably be expected to converge towards a consensus.
As such, we advocate an approach where the KR does not seek to reproduce and augment human expert consensus but rather positions itself as a \emph{tool for discovery} which, starting from purely objective nuggets of knowledge, enables and aids the human expert to investigate---and hopefully resolve---areas of apparent, or real, incoherence by comparing the dynamic consequences of various collections of independently-conceived rules.

In particular, our system does \emph{not} seek to figure out the `correct' necessary conditions for a nugget and even less seeks to impose a pre-conceived structure on the Kappa model implied by the contents of the KR: there is no specification of what the system `should be'; indeed, the basic philosophy of our approach can be summarized as: taking as input partial empirical knowledge of a system; and producing as output the various consequences of this knowledge, \eg the necessary conditions for PPIs, the causal structure, or pathways, that the system contains, \etc In a sense, from a computer science perspective, the workflow may seem backward: from partial knowledge of the behaviour of the system, we seek to \emph{determine} its specification. Our approach is thus intrinsically oriented towards \emph{systems}, not synthetic, biology which shares precisely this aim.

\paragraph{Overview of the paper.}
In section 2, we describe our graphical formalism, define our meta-model and formalize the notions of nuggets and their aggregation. In section 3, we illustrate these ideas with some simple examples. In section 4, we describe the instantiation process taking a given state of knowledge and chosen collection of proteins to a formal rule-based [Kappa] model. We conclude in section 5 with some remarks about our prototype implementation and directions for future work.

\section{The graphical formalism}

In this section, we first introduce a notion of (simple) graph with a second, independent (simple) graph structure on its nodes. This provides a more flexible and general starting point than in previous works \cite{danos2012graphs,danos2013constraining,danos2013thermodynamic} on `site graphs' for rule-based modelling where the permitted kinds of nodes and edges were hard-wired. We then introduce a particular graph---our \emph{meta-model}---which we use to \emph{type} graphs so as to reintroduce the previously hard-wired constraints in a transparent fashion.

\subsection{Structured graphs}

A \emph{structured graph} $G$ is defined by
\begin{itemize}
\item
a finite set of \emph{nodes} $\nodes$;
\item
two simple directed graph structures, $\mathcal{S}$ and $\edges$, \ie binary relations on $\nodes$;
\item
a function assigning, to each node $n \in \nodes$, a (possibly empty) set $V_n$ of possible values.
\end{itemize}
The graph structure $\mathcal{S}$ allows us to formalize the notion that a node may `belong to' another, \eg a node representing a region of some protein belongs to the node representing the protein in question. The graph structure $\edges$ represents links between nodes. Indeed, our definition has much in common with the essential intuition underlying bi-graphs although, for our purposes, the structuring relation $\mathcal{S}$ reflects a hierarchical organization of the nodes themselves, rather than the space in which they move. A node for which $V_n$ is non-empty represents some (fixed or variable) \emph{attribute} whose possible values are the elements of $V_n$, \eg the particular amino acid at a certain point in a protein sequence or the presence or absence of a post-translational modification like phosphorylation.

The various notions of `site graph' in the line of work on the definition of the rule-based modelling language Kappa in terms of graph rewriting \cite{danos2012graphs,danos2013constraining,danos2013thermodynamic} correspond to more constrained variants of this general definition. The principal present novelty is to avoid hard-wiring the various kinds of nodes that can exist---agents, sites, \etc---and their hierarchical structure. Instead, it proposes a homogeneous space of nodes with a structure $\mathcal{S}$ that can capture arbitrary relationships between nodes. Our particular choice of meta-model, made in section~\ref{Sec:MM}, assigns a DAG structure to $\mathcal{S}$ so that it does indeed correspond to a (generalized) notion of hierarchical organization.

\subsection{Homomorphisms}

A \emph{homomorphism} $\homo{h}{G}{G'}$ of structured graphs is a function $\tfun{h}{\nodes}{\nodes'}$ such that (i) $\mathcal{S}$-edges and $\edges$-edges are preserved; and (ii) values are preserved, \ie $V_n \subseteq V'_{h(n)}$ for all states $s$.

Structured graphs and their homomorphisms form a category $\SG$; a homomorphism is a mono if, and only if, its underlying node function is injective.
The category $\SG$ has all pull-backs, all push-outs and all pull-back complements over monos. As such, it possesses all the structure required to support general sesqui-push-out rewriting \cite{corradini2006sesqui}. The sub-category of monos has all multi-sums \cite{diers}.

Given a category $\mathcal{C}$ and an object $T$ of $\mathcal{C}$, the \emph{slice category} of $\mathcal{C}$ over $T$ has, as objects, all arrows $f:A \rightarrow T$ of $\mathcal{C}$ into $T$; and, as arrows from $f:A \rightarrow T$ to $f':A' \rightarrow T$, all arrows $h:A \rightarrow A'$ of $\mathcal{C}$ such that $f = f' \circ h$. Given a structured graph $T$, $\SG / T$ can be thought of as the category of structured graphs \emph{typed by} $T$. Standard categorical reasoning establishes that the slice category construction preserves the above categorical structure. Importantly, the `hierarchical' structure $\mathcal{S}_T$ of $T$ and its edges constrain the objects of $\SG / T$ so that the choice of base object $T$ reintroduces, in a flexible manner, the kinds of conditions and constraints that have been hard-wired in previously proposed definitions of site graphs.

In the next section, we make such a choice for our knowledge representation by defining a specific graph $\MM$ which types, in the slice category sense, all possible models that we wish to consider, \ie $\MM$ is our meta-model. In section~\ref{Sec:nuggets}, we introduce \emph{nuggets} as a particular class of the structured graphs that exist in $\SG / \MM$. However, the general procedure of defining a desired class of graphs by a choice of base object $T$ could be applied in many other situations; it provides a unifying framework for discussing a broad class of related, but distinct, graphical formalisms (including their sorting disciplines).

\subsection{The meta-model}\label{Sec:MM}

We now define the concrete structured graph $\MM$ that serves as our meta-model: it specifies the various kinds of nodes that can exist, the values---if any---they can take, defines a hierarchical structure $\mathcal{S}_{\!\MM}$ [the dotted arrows] on nodes and, finally, specifies the ways that edges can be placed [the solid arrows].

The names of nodes, such as `agent' and `BND', are not part of the formalism; they are a convenient labelling for the purposes of discussion.
Nodes with `jagged' outlines have a non-empty set of possible values, \ie are attributes: \texttt{flag} and \texttt{is\_bnd} are assigned the set $\set{0,1}$ of Booleans; \texttt{loc}, which identifies the position of a residue in a sequence, is assigned the set $\mathbf{Z^+}$ of positive integers; \texttt{aa}, which identifies amino acids, is assigned the standard 20-element set of one-letter amino acid codes, \ie all letters except B, J, O, U, X and Z; \texttt{int}, which specifies a sequence interval, uses  the set $\mathbf{Z^+} \times \mathbf{Z^+}$ of pairs $(\ell,h)$ of positive integers to specify sets of the form $\set{n \in \mathbf{Z^+} \st \ell \leq n \leq h}$ (so empty if $\ell > h$); and \texttt{bnd\_rc}, \texttt{brk\_rc} and \texttt{mod\_rc}, which specify the rate constants of actions, are assigned the set $\mathbf{R^+}$ of positive reals.
\[
\includegraphics[scale=0.61]{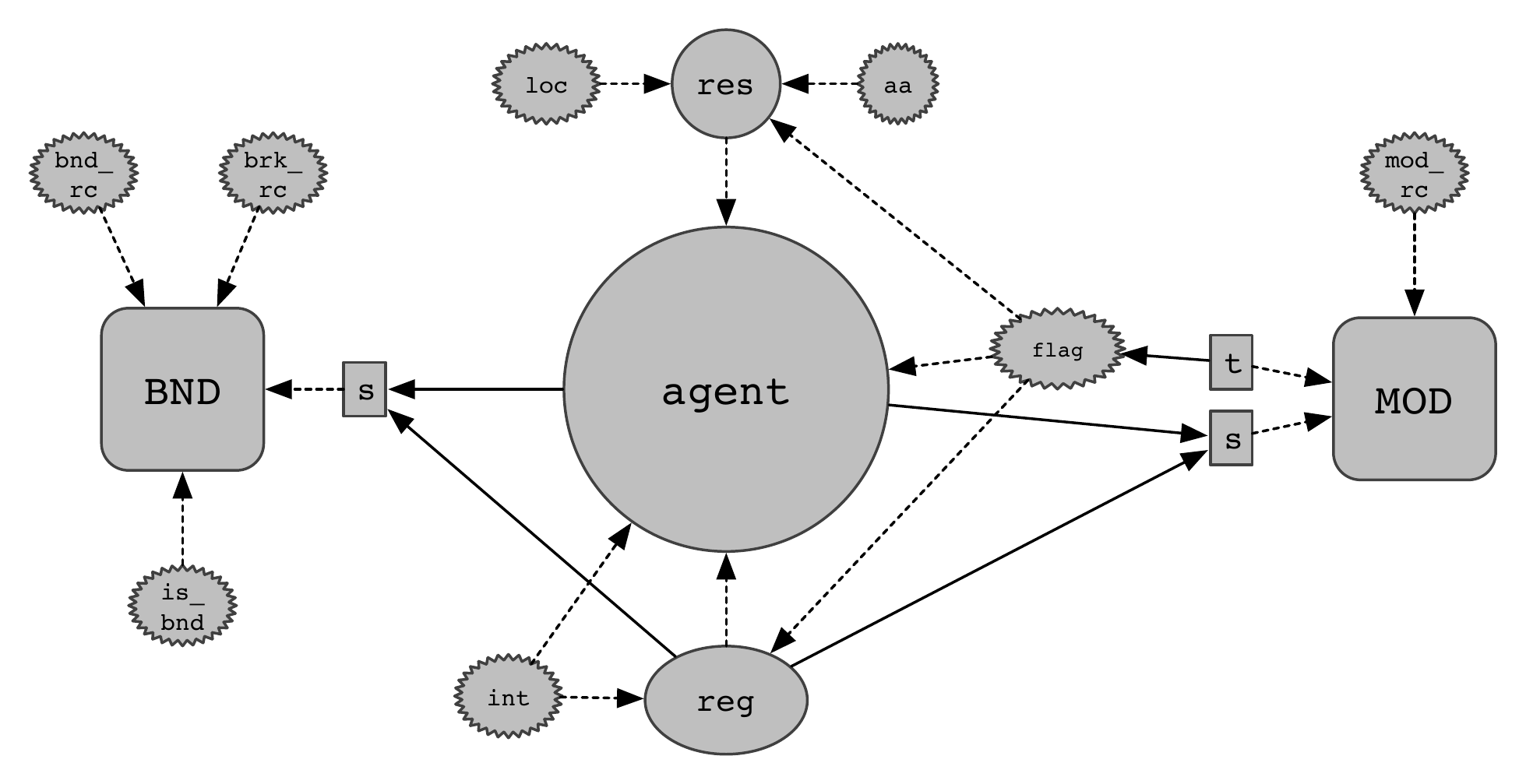}
\]

It is important to note that \texttt{MOD} actions only target flags. This reflects the fact that only flags represent `pieces of state' that can be modified \emph{within} the system and, as such, are reified in the eventual Kappa model; all other attributes are out of the system and, although they may play a r\^ole in the instantiation process leading to a Kappa model, will either not appear in the model at all or, alternatively, appear in such a way that they can only be tested but never modified.

A typing $\homo{h}{G}{\MM}$ formally designates each of the nodes of $G$ as being either an \emph{agent}, a \emph{reg(ion)}, a \emph{res(idue)}; or as one of the fixed attributes (\texttt{aa}, \texttt{loc}, \etc\!); or as a \emph{flag}; or as a \emph{binding} action (\texttt{BND}), \emph{modification} action (\texttt{MOD}) or as a \emph{s(ource)} for a \texttt{BND} or \texttt{MOD} or as a \emph{t(arget)} for a \texttt{MOD}. The mapping of $G$ into $\MM$ further implies that the graph structure $\mathcal{S}_G$ of $G$ must respect the restrictions imposed by $\MM$, \ie that regions and residues belong to agents; that nothing belongs to attributes or flags; and that sources and targets belong to their respective actions. Moreover, because $\mathcal{S}_{\!\MM}$ is a DAG, $\mathcal{S}_G$ must also be a DAG for any $G$ typed by $\MM$. Finally, the edge structure $\edges_G$ of $G$ must also respect the restrictions imposed by $\MM$; this means that only agents and regions can engage in binding actions; and that only flags can be targeted by modification actions.

This meta-model provides the foundation for a rigorous ontology for the kinds of information that are pertinent to rule-based descriptions of signalling networks, \ie proteins as agents, domains and other binding sites as regions, key amino acid locations as (key) residues, \etc However, the present framework remains purely formal and does not have any means to enforce correct semantic usage of this ontology; as such, we plan to augment our framework with a system of appropriate annotations in order to be able to carry out semantic checking and reasoning. We will return to this point later.

\subsection{Nuggets}\label{Sec:nuggets}

Let us now motivate the particular choice $\MM$ of meta-model made in the previous section by considering a typical `nugget' of knowledge in molecular biology: ``EGFR binds the SH2 domain of Grb2 provided that EGFR is phosphorylated and residue 90 of Grb2 is a serine". This would naturally be represented as the following structured graph [where dotted arrows are represented implicitly by nesting]:
\[\includegraphics[scale=0.65]{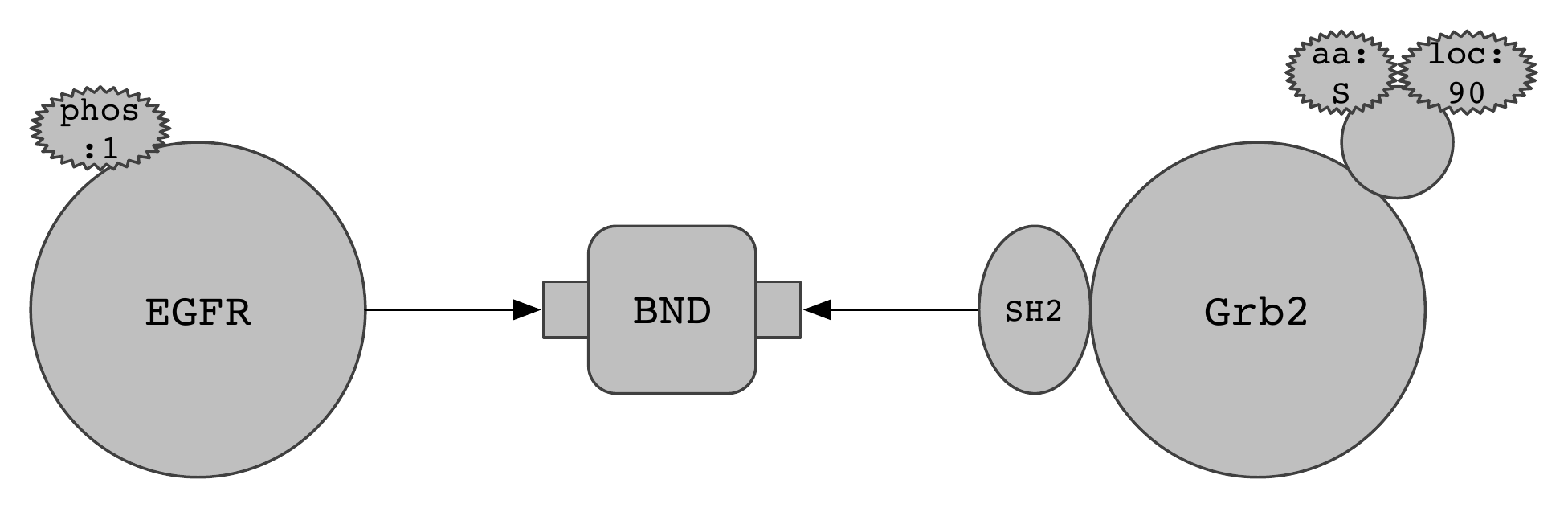}\]
where the agent `EGFR' has a state `phos' (with value 1, meaning true) and the agent `Grb2' has a region `SH2' and a residue (with unimportant name) located at position 90 of the sequence and which is required to be a serine (the value S). Note that no region has been specified on EGFR; the meta-model explicitly allows for this by the fact that a \texttt{BND} action can have either an agent or a region as its source.

Formally speaking, a \emph{nugget} is a connected graph $G$ typed by $\MM$ in such a way that each node has at most one copy of each of its possible attributes, where all attributes and flags have a uniquely specified value, the relation $\mathcal{S}_G$ is transitive-closed and exactly one action node does not have a specified value for its \texttt{is\_bnd} flag. This unique action is the \emph{principal} action of the nugget; any other (necessarily \texttt{BND}) actions represent the required presence or absence of their corresponding bonds in order for the principal action to be possible. We further ask that any \texttt{BND} has exactly two sources and any \texttt{MOD} has at most one source and one target (and at least one of the two). Other than asking for $G$ to be connected, all of these constraints are motivated by domain-specific considerations and, in due course, will be handled via semantic checking. 

Our representation language is thus a generalization of that typically used in rule-based modelling. The principal differences are that (i) the \emph{action} of a rule is represented explicitly as a node in the graph; (ii) binding actions can act directly on agents rather than necessarily via sites/regions; and (iii) static attributes and dynamic flags are represented as values associated to persistent nodes rather than as mutually exclusive sets of nodes. We draw a distinction between flags and attributes in order to make an explicit difference between things that can be modified `in the system', \eg the phosphorylation status of some residue, and things that can either only be modified `out of the system', \eg the identity of an amino acid that can only be modified by an external mutation event, or are pure meta-data, \eg UniProt IDs.

The latter points (ii) and (iii) are important in order to be able to represent biological knowledge as faithfully as possible: knowledge is often stated in a piecemeal and incomplete fashion but this should not prevent us from being able to formalize it. For example, when the site at one end of a bond is unknown, this can now be represented \emph{as is} with no need---as there would have been in standard rule-based modelling---to create a `fictitious' site. A second example could be the use of attributes to provide a transparent representation of detailed structural information, \eg about key residues of a protein or cases where a binding interaction depends on multiple PTMs that can otherwise only be opaquely encoded. We will return to point (i) in section~\ref{Sec:aggreg} where we will introduce the notion of nugget \emph{aggregation} which depends critically on the explicit representation of actions as nodes.

\subsection{Models}\label{Sec:models}

A site graph $\tfun{m}{M}{\MM}$ is a \emph{pre-model} of a collection $C$ of nuggets $\tfun{n_i}{N_i}{\MM}$ iff, for all $i$, the arrow $n_i$ factors through $m$. In words, $M$ is a graph, itself typed by the meta-model, which types all of the nuggets; indeed, $M$ can be thought of as a summary statement of the collection $C$ of nuggets. We refer to the pair $(C,M)$ as a \emph{model}.

Note that a given collection of nuggets may be assigned many different pre-models and that this choice affects how those nuggets are to be interpreted: the import of any particular choice is that it identifies which nodes in one nugget correspond to those in another. In the above example, we have a node labelled as `EGFR'; but that label does not exist in the formalism so, if we have a second nugget which also speaks of the same agent `EGFR', we need a way to say that these two nodes are the same. The pre-model gives us precisely this possibility: the two agents are mapped to the same node of $M$, \ie nodes of $M$ provide labels/names for the nodes of nuggets. This means that two different pre-models, $M_1$ and $M_2$, for the same collection $C$ of nuggets can have completely different meanings; in particular, $C$ and $\MM$ provide two (uninteresting) extremes with interesting cases lying in between. In general, the two components of a model evolve together as we add more and more information; we discuss this briefly in section~\ref{Sec:aggreg}.

Let us note here that the necessity of a pre-model partially arises in order to enforce minimal semantic coherence in our formal framework. If we had semantic annotations that uniquely identify agents, we could potentially use them---instead of a pre-model---to solve the above cross-nugget identification problem. However, we have chosen to take a different approach so as to provide a more flexible notion of agent: in a general site graph---and, in particular, in a pre-model---the \texttt{aa} attribute of a residue may be assigned a \emph{set} of one-letter codes in order to express the fact that an agent represents, in general, a \emph{neighbourhood} in sequence space rather than a unique sequence. This flexibility affords us the possibility of organizing knowledge about minor variants of a protein using a single agent; this (rather prosaically) pre-empts the need to define, and name, lots of tedious variants but, more to the point, matches everyday practice in biology where, for example, (wild-type) `Ras${}_{WT}$' and (mutant) `Ras${}_{G12V}$' are both thought of as being `Ras'---they just differ in one or two small, although possibly very significant, ways.

This notion of model is therefore our first step towards a full semantic layer for our knowledge representation scheme: semantic annotations will be made at the level of pre-models, not individual nuggets, so as to minimize the amount of needed annotation and, more importantly, to ease the maintenance of semantic coherence across the entire current collection of nuggets. It should be noted that this approach to grounding differs from more traditional approaches, such as that used by BioPAX \cite{demir2010biopax}, which insist upon each formal entity corresponding to a unique physical entity. Indeed, our approach is, by design, particularly attuned to the representation of signalling networks and, as such, is less constrained than BioPAX which, as a framework of far broader applicability, has to bear a far stronger semantic burden.

\subsection{Aggregation}\label{Sec:aggreg}

Suppose that, at some point in time, we have a model $(C,M)$ and that we now obtain a new nugget $N'$. We always have the possibility simply to add $N'$, yielding a new collection $C'$ of nuggets; this might necessitate updating the underlying pre-model $M$ to $M'$ in the event that $N'$ contains entirely novel nodes or edges. A pre-model update is specified by a co-span, or glueing, $N' \rightarrow M' \leftarrow M$ from the multi-sum of $N'$ and $M$ that specifies exactly which nodes and edges of $N'$ already existed in $M$ and which have been added between $M$ and $M'$.

It is important to note that the identification of which nodes of $N'$ already existed in $M$ includes action nodes; this means that the co-span may identify the action of $N'$ as being the \emph{same} as that of some pre-existing $N$ in $C$. Alternatively, the action of $N'$ may be newly added; this might result in $(C',M')$ having two distinct actions involving (some of) the same agents, a situation that may or may not be desirable. Sometimes, for example, two proteins can indeed bind each other in two distinct ways; however, it could also be the case that $N'$ has actually brought some new information about a single binding interaction that we would rather use to \emph{update} the pre-existing nugget $N$. Such an update amounts to the assumption that the nuggets $N$ and $N'$ represent the same interaction \emph{mechanism}; but this does not necessarily mean that the two nuggets refer to exactly the same agents since mechanisms can be shared across families of proteins. We return to this point in section~\ref{Sec:exagg} after describing the formal process of \emph{aggregation}.

If we wish to \emph{update} the nugget $N$ with the information contained in $N'$, we need to specify two things: the new information brought by $N'$ and any \emph{deprecated} information in $N$ that should now be removed. The former is specified by the choice of a co-span of monos $h_+ : N \rightarrow N_+ \leftarrow N' : h'_+$ from the multi-sum of $N$ and $N'$; while the latter is specified by a mono $\homo{h_-}{N_-}{N}$ where $N'$ is that part of $N$ that we wish to preserve. In most cases, there is a canonical choice of co-span given by the intuitive unification of $N$ and $N'$ but, in cases where there are non-trivial automorphisms of $N$ and $N'$, the unification process may be non-deterministic and a choice becomes necessary.

The pull-back complement of $h_-$ and $h_+$ defines a graph $N_\pm$ containing precisely the new information from $N'$ with all deprecated information from $N$ removed. Formally, this is exactly a step of graph rewriting taking $N$ to $N_\pm$. In the event that nothing is to be removed from $N$, \ie $h_-$ is the identity on $N$, this rewriting step degenerates to being simply the refinement of $N$ to $N_+$ as specified by $h_+$.
\[\xymatrix@C=10pt@R=10pt{
& N \ar[dr]^{h_+}  \ar@{~>}[dd] & & N' \ar[dl]^{h'_+} \\
N_- \ar[ur]^{h_-} \ar@{.>}[dr] \pullbackcorner[r] & & N_+ \\
& N_\pm \ar@{.>}[ur]
}\]
This step of rewriting is also propagated to the pre-model $M$, resulting in a new model $(C',M')$ where, unlike in the case of adding $N'$, $C'$ clearly has the same number of nuggets as $C$.

\section{Examples of aggregation}\label{Sec:exagg}

\paragraph{Nugget update}
Consider updating the example nugget $N$ of section~\ref{Sec:nuggets} with the information contained in $N'$:
\[\includegraphics[scale=0.65]{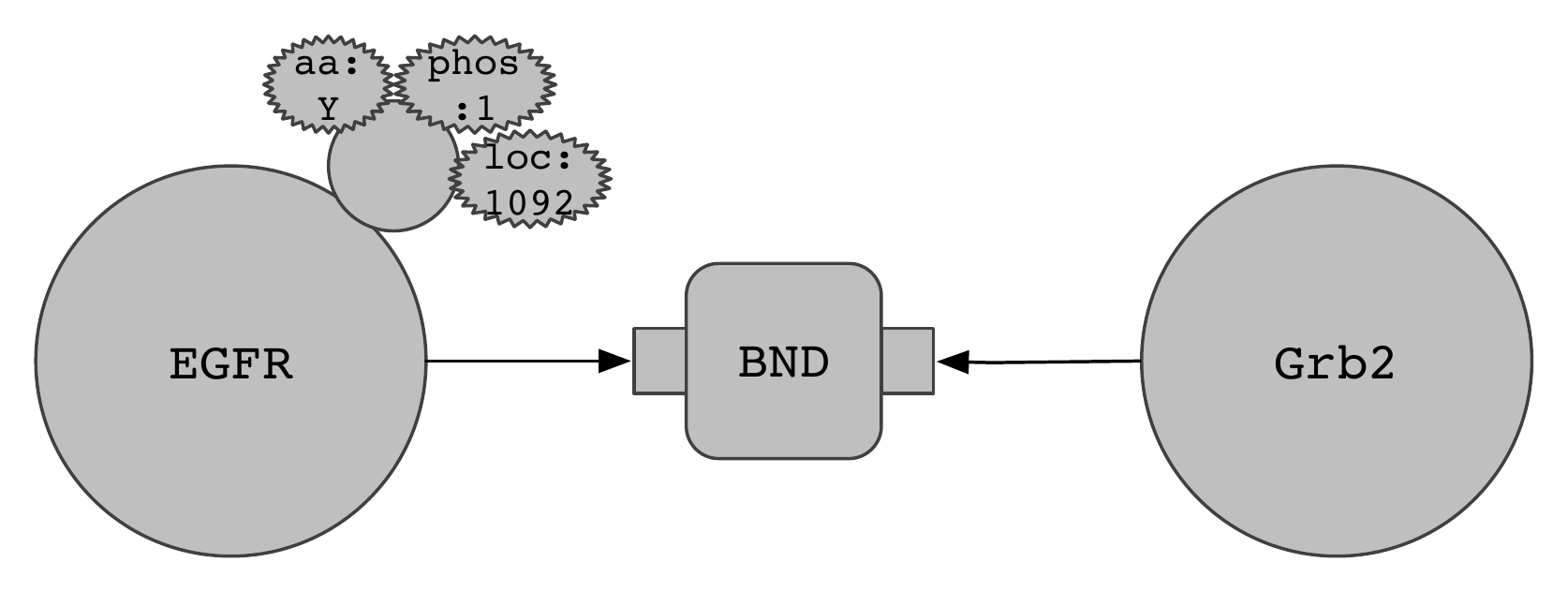}\]
\ie ``EGFR binds Grb2 provided EGFR is phosphorylated on Y1092".

If we choose not to deprecate anything from the original nugget, we obtain:
\[\includegraphics[scale=0.65]{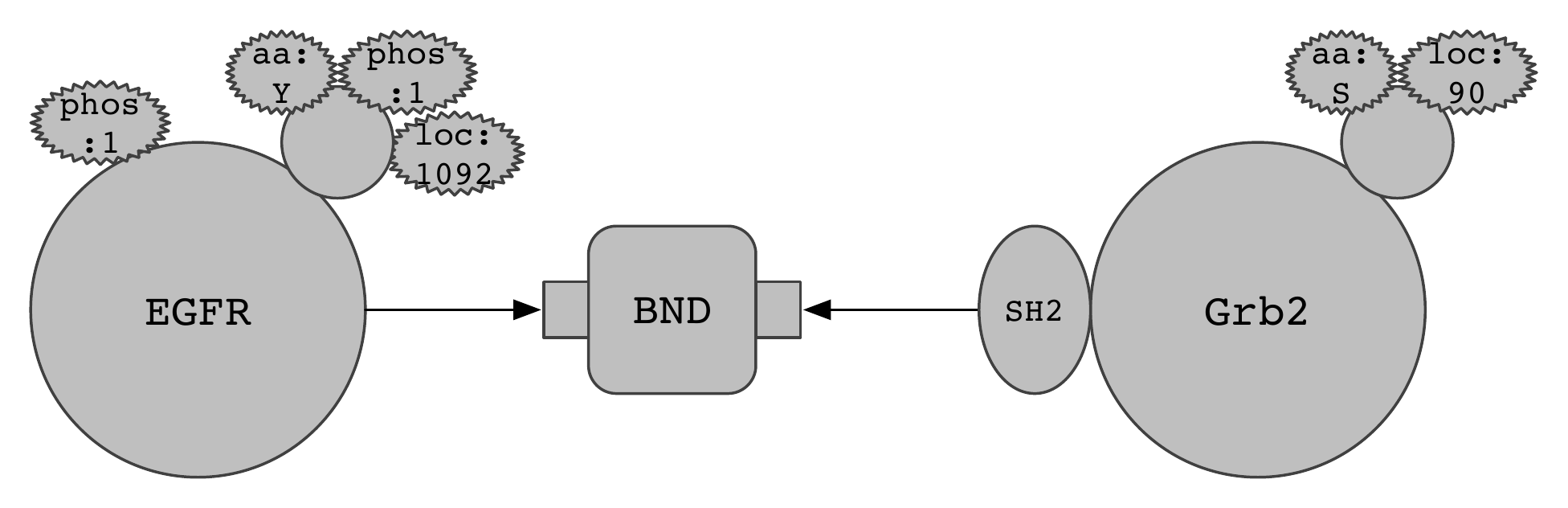}\]
If instead we were to specify that the \texttt{phos} flag from $N$ is to be removed, we would obtain:
\[\includegraphics[scale=0.65]{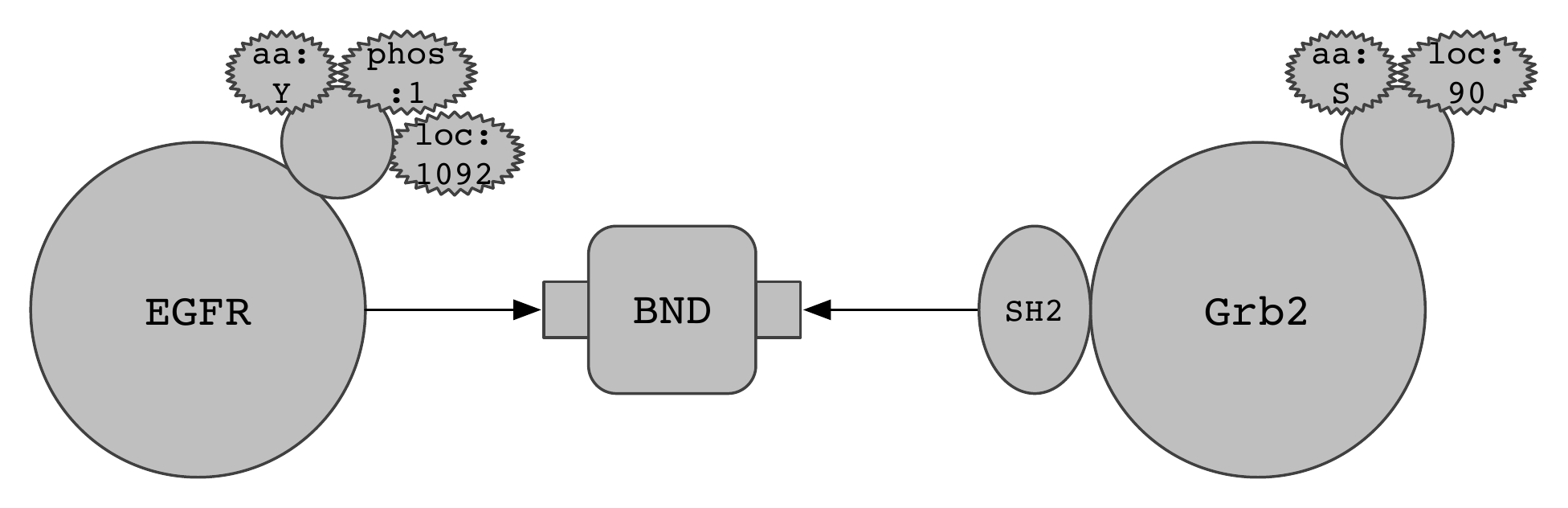}\]

We also have the possibility of specifying that the \texttt{phos} flag of \texttt{EGFR} is to be \emph{moved} to the residue at \texttt{loc}ation 1092; this slightly different rewrite would result in the same revised nugget (as immediately above) but would additionally maintain, in the pre-model, any \texttt{MOD} actions that were already known to act upon the flag. Such a rewrite reflects a common model update situation that arises when finer-grained empirical data allows a more precise assignment of `pieces of state' to a protein's constituent parts. This possibility is an immediate consequence of the built-in flexibility provided by the DAG structure of the meta-model and also depends on the transitive-closure, with respect to that DAG, required of nuggets.

\paragraph{Nugget aggregation}
Unification can also be partial: if we further update the previous nugget with
\[\includegraphics[scale=0.65]{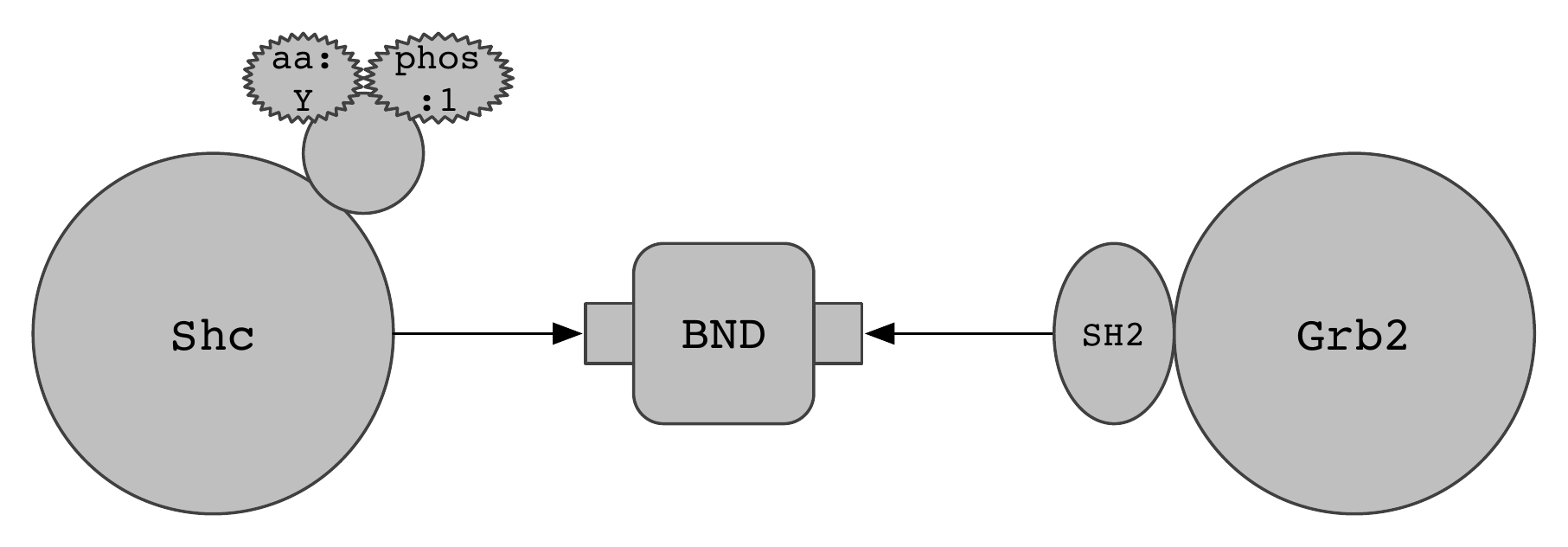}\]
\ie ``tyrosine-phosphorylated Shc binds the SH2 domain of Grb2", by unifying the \texttt{Grb2} and the \texttt{BND} nodes, we obtain:
\[\includegraphics[scale=0.65]{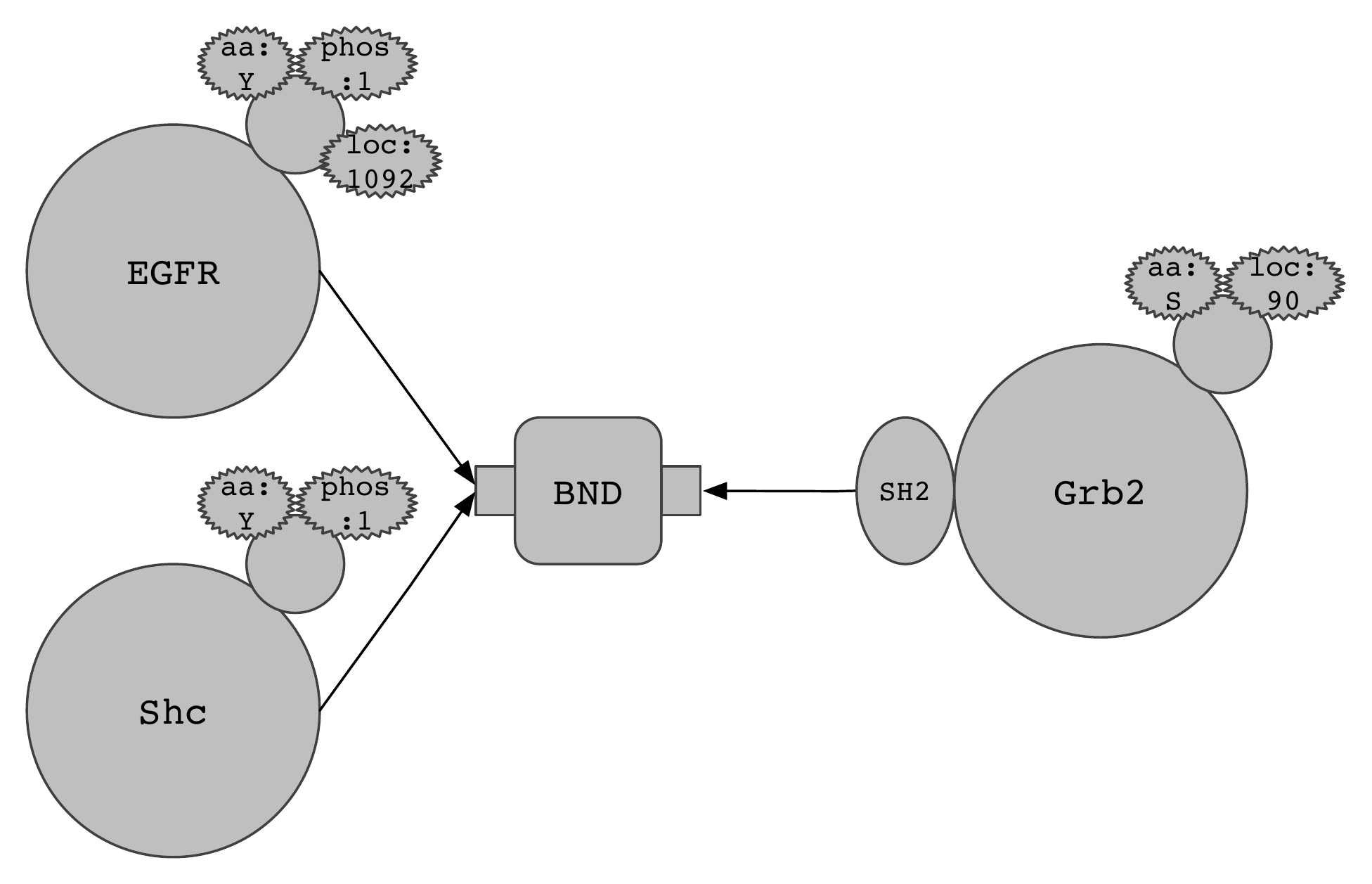}\]

Note how the contextual conditions on \texttt{Grb2} that occur in the original nugget are propagated, by the very process of aggregation, to its newly added interaction with \texttt{Shc}. This is a typical example of the use of our framework as a tool for discovery---at the level of necessary conditions for PPIs---which renders completely transparent the `by similarity' style of reasoning which is ubiquitous in molecular biology.

Note also that, by fusing the two \texttt{BND} nodes, we force \texttt{EGFR} and \texttt{Shc} to target the same source of the (unique) \texttt{BND} node: this nugget has a disjunctive interpretation, giving rise to two distinct Kappa rules, and should be read, modulo the contextual conditions, as ``EGFR or Shc binds the SH2 domain of Grb2". The importance of this is that, in the translation to Kappa, agents will be assigned one \emph{site} for each \texttt{BND} action in which they participate: had we unified only the \texttt{Grb2} nodes, leaving the two \texttt{BND} nodes distinct, \texttt{Grb2} would have been given two sites, one to bind \texttt{EGFR} and the other for \texttt{Shc}, with no conflict between the two generated Kappa rules; however, after further merging the \texttt{BND} nodes, it would receive only one site, giving rise to an intrinsic conflict between the two generated rules. Let us now consider this \emph{instantiation} process in detail.

\section{Model instantiation}

In this section, we illustrate how and why formal sites are reified, starting from our running example in the KR, leading to a strict rule-based model that reflects exactly and only the constraints warranted by the knowledge in the KR. This reflects our general philosophy that, in the absence of information to the contrary, we should always draw the most general conclusions possible: so two \texttt{BND} actions should be considered independent unless we have evidence to the contrary. In this way, we achieve a pragmatically (and cognitively) convenient separation of concerns whereby knowledge is integrated and aggregated as we learn it; while its eventual consequences for a future Kappa model are determined, at model generation time, by the fully automatic instantiation procedure that produces an executable Kappa model.

\subsection{Agent variants}

Consider the following pre-model which types the various example nuggets from the previous section:
\[\includegraphics[scale=0.65]{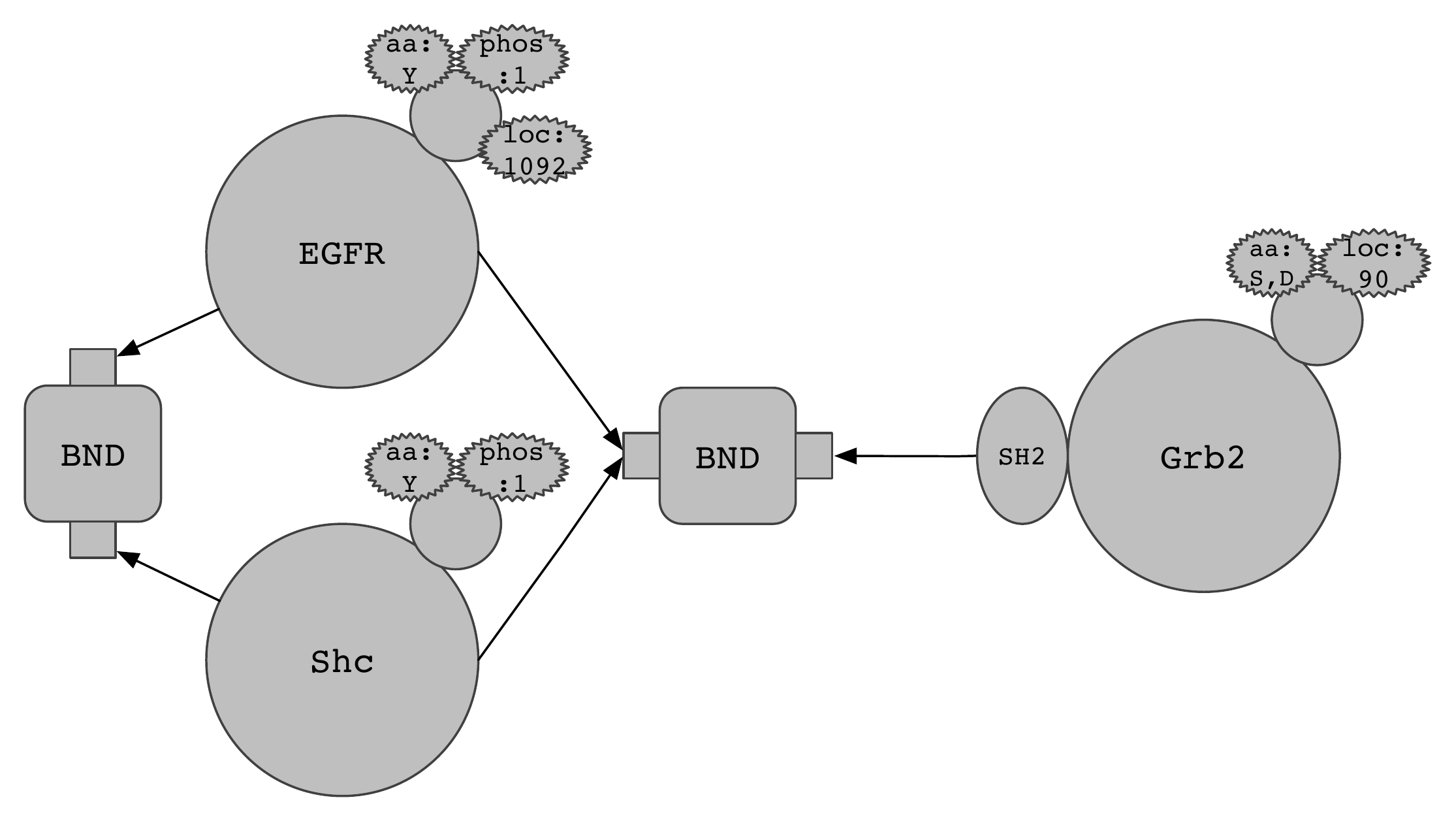}\]
The agent \texttt{Grb2} has a choice of amino acids: at \texttt{loc}ation 90, it can be S [serine] or D [aspartic acid]. Moreover, the \texttt{BND} action between \texttt{Grb2} and \texttt{EGFR} or \texttt{Shc} depends on this: the action can only occur if serine, not aspartic acid, is present. As such, this residue must be reified as a formal site in Kappa such that it has two possible internal states, S and D, which are never modified by any rule in the system.

\subsection{Site conflict invariants}

There are two ways in which \texttt{BND} actions give rise to conflicts. Firstly, in the case of an aggregated nugget, there is---by hypothesis---an \emph{intrinsic} conflict between the disjuncts; this reflects the fact that we consider them all to use the same \emph{mechanism} to bind a partner and, as such, they all compete for the same binding site on that partner. Secondly, two \texttt{BND} actions may have specified sequence \texttt{int}ervals, defining their physical footprints, which have an overlap; in this case, the binding mechanisms differ but, for the obvious reason that two binding partners cannot occupy the same space at the same time, there is an \emph{extrinsic} conflict (also known as `steric occlusion').

In the first case, we need to reify a single formal site for which all disjuncts compete. In the second case, the two overlapping regions should still, in general, be reified as two distinct formal sites; however, we must also keep track of the fact that these sites are in conflict. Note, however, that this conflict relation is symmetric, but not transitive, in general; as such, we cannot necessarily merge two conflicting sites as one, but not the other, may also be in conflict with a third site. On the other hand, any \emph{clique} of conflicting sites can be conflated to a single formal site in a subsequent optimization phase.

In our pre-model, there is an intrinsic conflict between \texttt{EGFR} and \texttt{Shc} to bind \texttt{Grb2}. However, there is no extrinsic conflict between \texttt{Grb2} and \texttt{Shc} to bind to \texttt{EGFR} because, in the absence of overlapping interval information, we assume the two binding actions to be independent. Overall, we therefore obtain the following strict rule-based model:
\[\includegraphics[scale=0.65]{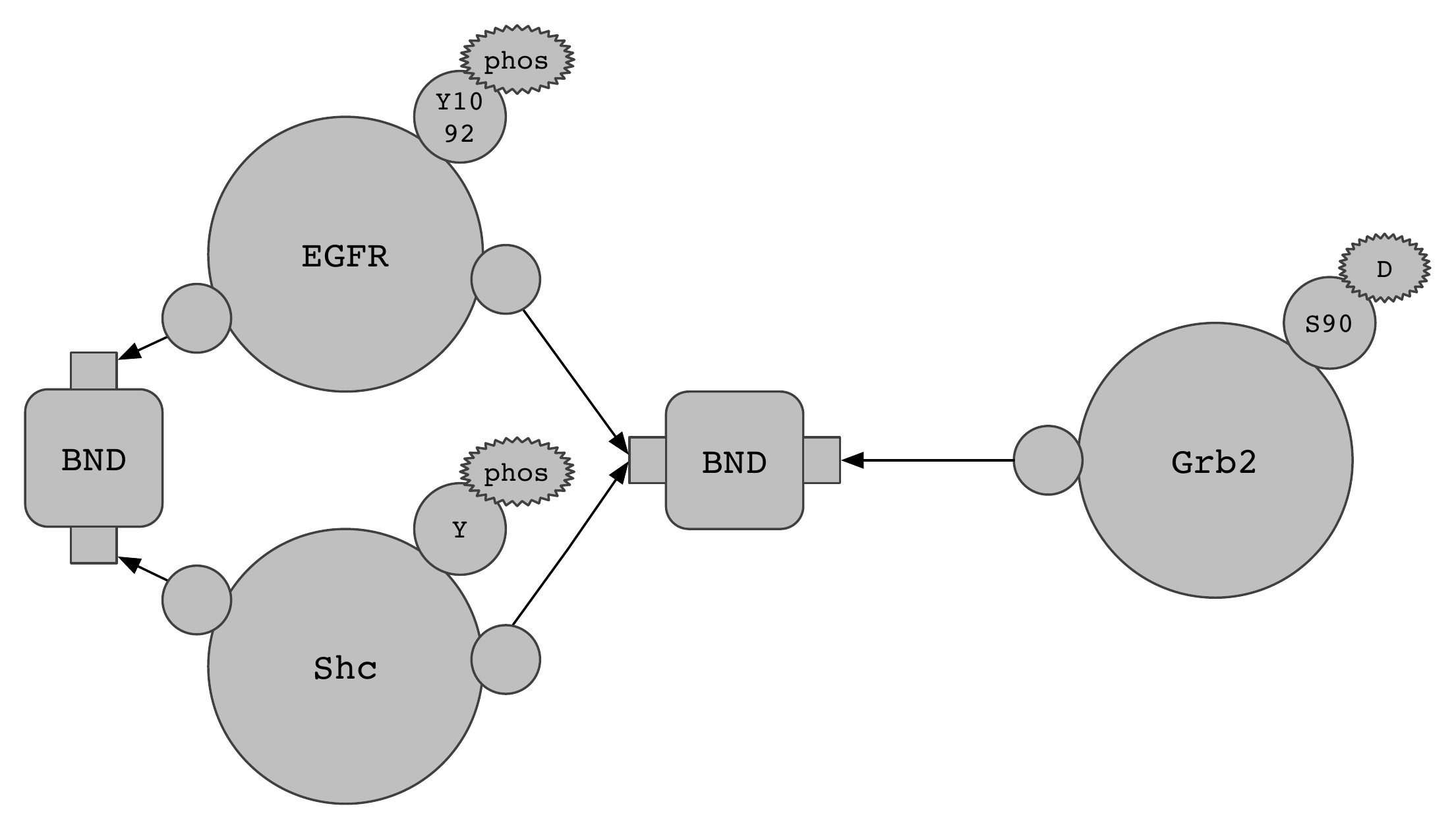}\]
Note that all binding actions now include (formal) sites; and that residues have also become (formal) sites with single pieces of (binary or, more generally, $n$-ary) state: in the case of \texttt{Grb2}, the residue at \texttt{loc}ation 90 is considered to be in state S by default---but which can be overridden to state D to represent the mutated protein. We refer to the actions in this graph as \emph{pre-rules}; they differ from standard Kappa rules only in the possibility, as shown here, of retaining intrinsic conflicts as disjunctive rules.

\subsection{Kappa}

The last phase of the automatic instantiation procedure effects the final compilation into executable Kappa. It specifies all agent signatures, `multiplies out' disjunctive rules and provides a (context-less) unbinding rule for each binding action. 

The agent signatures are:
\begin{verbatim}
%agent: EGFR(bShc,bGrb2)
%agent: Shc(bEGFR,bGrb2)
%agent: Grb2(rgSH2,rs90~S~D)
\end{verbatim}
The state of residue \texttt{rs90} is specified, as discussed above, to be S by default. Note also that the sites \texttt{bGrb2} of \texttt{EGFR} and \texttt{Shc} are distinct sites: they have the same name but still belong to different agents.

The three binding actions are translated into Kappa rules as follows. The intrinsic conflict between \texttt{EGFR} and \texttt{Shc} is captured by their competition for the \texttt{rgSH2} site of \texttt{Grb2}:
\begin{verbatim}
EGFR(bGrb2),Grb2(rgSH2,rs90~S) -> EGFR(bGrb2!0),Grb2(rgSH2!0,rs90~S)
EGFR(bGrb2!0),Grb2(rgSH2!0) -> EGFR(bGrb2),Grb2(rgSH2)

Shc(bGrb2),Grb2(rgSH2,rs90~S) -> Shc(bGrb2!0),Grb2(rgSH2!0,rs90~S)
Shc(bGrb2!0),Grb2(rgSH2!0) -> Shc(bGrb2),Grb2(rgSH2)

EGFR(bShc),Shc(bEGFR) -> EGFR(bShc!0),Shc(bEGFR!0)
EGFR(bShc!0),Shc(bEGFR!0) -> EGFR(bShc),Shc(bEGFR)
\end{verbatim}

In the case that we had introduced a sequence \texttt{int}erval overlap between the two binding actions concerning \texttt{EGFR}, we would acquire a site conflict requiring us to maintain the invariant that at most one of \texttt{bShc} and \texttt{bGrb2} can be bound at any given time:
\begin{verbatim}
EGFR(bShc,bGrb2),Shc(bEGFR) -> EGFR(bShc!0,bGrb2),Shc(bEGFR!0)
EGFR(bShc,bGrb2),Grb2(rgSH2,rs90~S) -> EGFR(bShc,bGrb2!0),Grb2(rgSH2!0,rs90~S)
\end{verbatim}

\section{Conclusion}

We have presented a meta-model for the representation of the kind of knowledge required to build rule-based models of cellular signalling networks in which we can transparently include mutational variants of proteins in order to investigate their dynamic consequences. The framework provides a generalization of the usual strict rule-based meta-model and, in particular, represents the actions of rules explicitly as nodes. This enables the key notion of aggregation which serves as our source of biologically-plausible inference. The presentation is entirely mathematical, being framed in terms of a graph rewriting formalism, but can be considered as a specification for an actual system. We are currently building a prototype implementation of the framework, including the automatic translation of a collection of nuggets into \emph{bona fide} Kappa, whose implementation details will be discussed in a forthcoming paper.

The instantiation process guarantees that (i) each concrete agent that is generated must have a unique value for each of its \texttt{aa} attributes, manifesting as \emph{unmodifiable} states in the resulting Kappa; and (ii) intrinsic and extrinsic conflicts between sites are propagated to the formal Kappa model. The first point implies that a rule testing for a wild-type value of such an attribute would only apply to a concrete agent that is not mutated at that residue; conversely, a rule that tests for a non-wild-type value would not apply to a concrete agent unless it had undergone an appropriate mutation. This enables a transparent account of both loss- and gain-of-function mutations that side-steps the various difficulties (concerning gain-of-function mutations) that were encountered by the original `meta Kappa' project \cite{danos2009rule,harmer2009rule}. The second point relieves the modeller from having to keep track of all conflict invariants that must be maintained and separates the distinct concerns of the open world of the KR and the closed world of a particular model.

Finally, the framework presented here remains entirely formal and, although we obviously have in mind an interpretation in terms of signalling networks, it does not actually embody any domain-specific knowledge. In particular, although we have used suggestive names for nodes in our examples, these have no actual significance and could be arbitrarily renamed without affecting the content of the knowledge representation or any generated Kappa models. We plan to address this issue this by introducing semantic annotations for nodes---as pure meta-data attributes---that would allow us to express domain-specific properties such as ``this region is an SH2 domain" or ``this residue must be serine or threonine" from which \emph{automatic rewrites}, expressing update and/or aggregation of nuggets, could be determined.

This \emph{grounding} process opens up the possibility of performing a second level of semantic checking on nuggets that have already been verified to be syntactically well-formed. Such checks could be purely routine, \eg ``binding actions have exactly two participants", ``only a kinase can phosphorylate a protein" or ``a serine/threonine kinase cannot phosphorylate a tyrosine residue". However, we principally envisage semantic annotations as a means to perform \emph{automatic rewriting} so as to make certain default aggregation decisions, \eg ``an SH2 domain can bind only one phospho-tyrosine ligand at a time". As such, we envisage a particular focus on the numerous domain-domain and domain-ligand PPIs that occur in signalling networks which, by their very nature, embody highly generic binding mechanisms constrained by relatively simple---but tedious and error-prone to write by-hand---intrinsic conflict conditions.

\paragraph{Acknowledgement.}
This work was sponsored by the Defense Advanced Research Projects Agency (DARPA) and the U. S. Army Research Office under grant numbers W911NF-14-1-0367 and W911NF-15-1-0544. The views, opinions, and/or findings contained in this report are those of the authors and should not be interpreted as representing the official views or policies, either expressed or implied, of the Defense Advanced Research Projects Agency or the Department of Defense.

\bibliographystyle{eptcs}
\bibliography{harmer}

\end{document}